\pdfoutput=1
\documentclass[11pt]{article}

\usepackage[preprint]{acl}
\usepackage{times}
\usepackage{latexsym}

\usepackage[T1]{fontenc}
\usepackage[utf8]{inputenc}

\usepackage{microtype}

\usepackage{inconsolata}

\usepackage{subfig}
\usepackage{amsmath}
\usepackage{amssymb}
\usepackage{multirow}
\usepackage{booktabs}
\usepackage{makecell}

\usepackage{graphicx}
\graphicspath{{figures/}}

\usepackage{threeparttable}
\newcommand{\squishlist}{
	\begin{list}{$\bullet$}
		{ \setlength{\itemsep}{0pt}
			\setlength{\parsep}{1pt}
			\setlength{\topsep}{1pt}
			\setlength{\partopsep}{0pt}
			\setlength{\leftmargin}{0.8em}
			\setlength{\labelwidth}{1em}
			\setlength{\labelsep}{0.5em} } 
	}
	
	\newcommand{\squishend}{
	\end{list}  
}

\title{EFPC: Towards Efficient and Flexible Prompt Compression}

\author{
  \textbf{Yun-Hao Cao},
  \textbf{Yangsong Wang},
  \textbf{Shuzheng Hao},
  \textbf{Zhenxing Li},
  \textbf{Chengjun Zhan},
  \textbf{Sichao Liu},
  \textbf{Yi-Qi Hu}
\\
   Huawei Technologies
\\
  \small{
  	\tt{caoyunhao@huawei.com}, \tt{wangyangsong1@h-partners.com},
     \tt{\{haoshuzheng1,lizhenxing11\}@huawei.com}, 
  }
\\
  \small{
  \tt{zhanchengjun@hisilicon.com}, \tt{\{owen.liusichao,huyiqi2\}@huawei.com}
}
}

\begin{document}
\maketitle
\begin{abstract}
	The emergence of large language models (LLMs) like GPT-4 has revolutionized natural language processing (NLP), enabling diverse, complex tasks. However, extensive token counts lead to high computational and financial burdens. To address this, we propose Efficient and Flexible Prompt Compression (EFPC), a novel method unifying task-aware and task-agnostic compression for a favorable accuracy-efficiency trade-off. EFPC uses GPT-4 to generate compressed prompts and integrates them with original prompts for training. During training and inference, we selectively prepend user instructions and compress prompts based on predicted probabilities. EFPC is highly data-efficient, achieving significant performance with minimal data. Compared to the state-of-the-art method LLMLingua-2, EFPC achieves a 4.8\% relative improvement in F1-score with 1\% additional data at a 4× compression rate, and an 11.4\% gain with 10\% additional data on the LongBench single-doc QA benchmark. EFPC's unified framework supports broad applicability and enhances performance across various models, tasks, and domains, offering a practical advancement in NLP.
\end{abstract}

\section{Introduction}

The rise of large language models (LLMs) such as GPT-4 has significantly advanced the field of natural language processing, making it possible to tackle a wide range of complex tasks. Various prompting techniques, such as Chain-of-Thought (COT)~\cite{CoT:NIPS22}, In-context Learning (ICL)~\cite{ICL}, and Retrieval Augmented Generation (RAG)~\cite{RAG:NIPS20}, have been instrumental in maximizing the potential of these models by generating rich and informative prompts. However, these methods often require prompts that are tens of thousands of tokens long, resulting in increased computational and financial overhead as well as diminished information perception abilities of the LLMs (e.g., degraded performance when processing noisy and lengthy contexts~\cite{Longllmlingua}).

\begin{figure}
	\centering
	\includegraphics[width=0.95\linewidth]{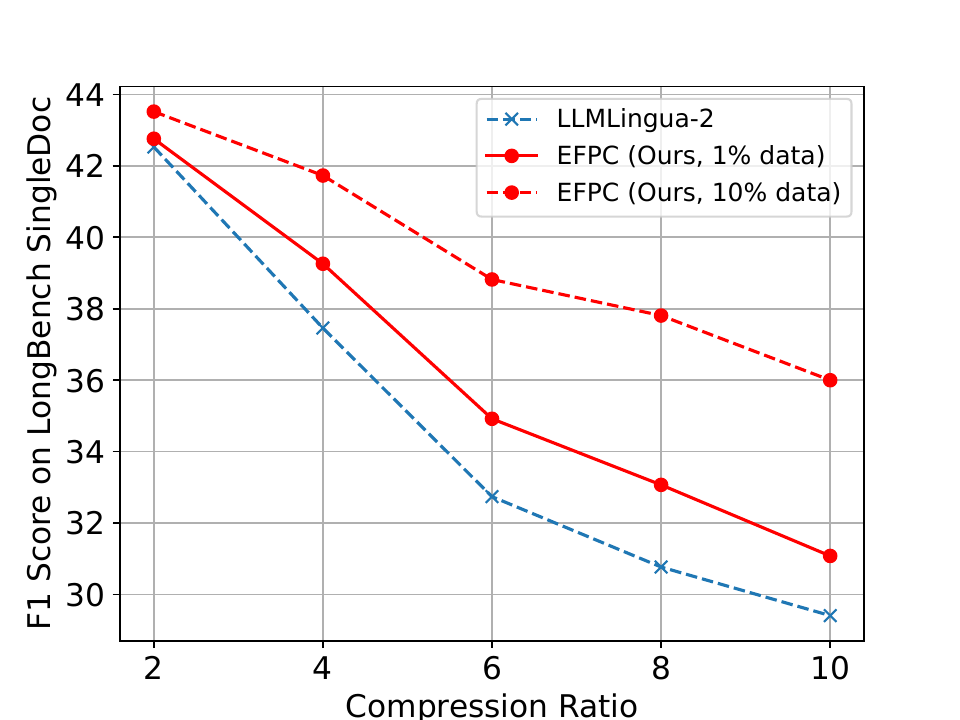}
	\caption{Performance under varying compression rates and training data amounts. Our EFPC method achieves significant improvements with minimal training data and larger gains with higher compression rates.}
	\label{fig:performance}
\end{figure}

To address these challenges, prompt compression has emerged as a promising solution. The goal is to shorten prompts without losing essential information, thereby improving efficiency and reducing costs. Existing methods generally fall into two categories: task-aware and task-agnostic. Task-aware methods~\cite{Longllmlingua, recomp:ICLR24, discreteCompress, Boostllm}, tailor compressed prompts to specific tasks or queries, achieving enhanced performance on downstream tasks. However, these methods often face challenges in terms of efficiency and generalizability, requiring multiple compressions of the same document depending on the associated queries and typically relying on time-consuming Transformer decoder architectures~\cite{transformer:Vaswani:NIPS17}. Task-agnostic methods~\cite{llmlingua, compresscontext}, propose to compress prompts by removing tokens or lexical units based on information entropy, irrespective of the downstream task. Recently, LLMLingua-2~\cite{llmlingua2} approachs prompt compression as a token
classification task and take the predicted probability of each token
being labeled as preserve as the compression metric. While offering better generalizability, these methods suffer from insuperior performanace on tasks such as question answering.

Building on previous foundational works, we propose an Efficient and Flexible Prompt Compression method, named EFPC, which can adapt between task-aware and task-agnostic scenarios based on the specific use case to maintain high efficiency. Specifically, our method leverages GPT-4 to compress prompts while ensuring their ability to perform tasks such as answering user questions. These compressed prompts are then paired with the original ones to create a binary classification training set, and a binary classifier with a Transformer encoder architecture~\cite{bert:devlin:NAACL19} is trained on this set. During training and inference, user instructions are concatenated with the original prompts to predict the retention probability for each word, which is then used to compress the prompts by discarding low-probability words. For task-agnostic compression, the user instruction is set to an empty string.

Moreover, this work addresses two primary challenges in the field: balancing efficiency and accuracy, and mitigating high data costs associated with prompt compression.

\squishlist
	\item{Efficiency and Accuracy Balance}. Existing work such as LongLLMLingua~\cite{Longllmlingua} employs decoder-based LLM LLaMA-2-7B~\cite{llama2} for prompt compression, achieving high accuracy at the expense of efficiency, making it less suitable for resource-limited scenarios requiring low latency. In contrast, encoder-based methods like LLMLingua-2~\cite{llmlingua2} offer a more lightweight structure that enhances efficiency but underperforms on tasks like question answering (QA). Our method bridges these gaps by enabling LLMLingua-2 to incorporate both task-aware and task-agnostic capabilities, significantly improving performance in QA and similar tasks.
	
	\item{Data Efficiency}. LLMLingua-2 uses data distillation with GPT-4 to create compressed datasets, incurring high financial and energy costs. Our method enhances data efficiency within the same framework as~\citet{llmlingua2}. As illustrated in Figure~\ref{fig:performance}, our method achieves a 4.8\% improvement in F1-score on the LongBench~\cite{longbench} single-doc benchmark with only 1\% more data at a 4x compression rate. When the additional data is slightly increased to 10\%, the relative improvement markedly rises to 11.4\%. Our approach boosts performance while minimizing additional data use, cutting API costs and energy consumption.
\squishend

In summary, our EFPC offers an innovative and efficient solution to prompt compression, contributing to the practical advancements in NLP. The primary contributions of our work are as follows:

\squishlist
	\item Flexible Compression Framework: EFPC effectively bridges task-aware and task-agnostic approaches through a simple switch in the prompt prefix. This dual capability substantially improves the performance of tasks like document question answering.
	\item Data Efficiency: Our method is highly data-efficient and achieves significant improvements with minimal additional data, reducing the need for costly API usage and thereby saving both money and energy.
	\item Training and Inference Efficiency: Unlike traditional decoder-based methods, EFPC employs a lightweight encoder structure, leading to more efficient training and inference processes. 
	\item Generalizability. We validated the effectiveness of our approach across various tasks and domains using multiple benchmarks. Moreover, our method is compatible with different compression models and target LLMs. It is designed for plug-and-play integration, demonstrating excellent generalization capabilities.
	
\squishend

\section{Related Works}

\begin{figure}
	\centering
	\includegraphics[width=0.9\linewidth]{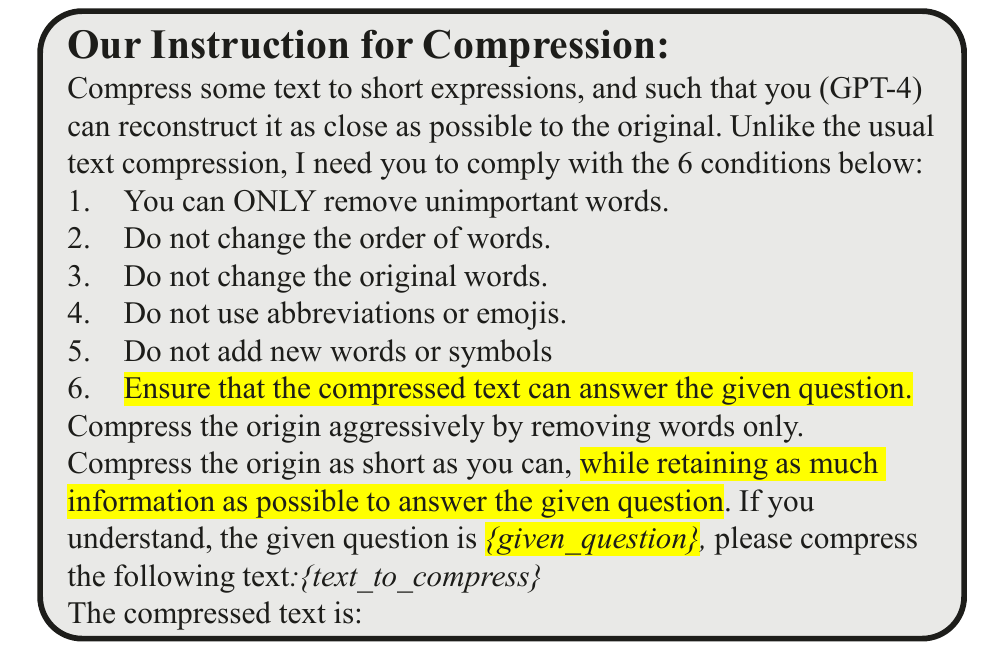}
	\caption{Our instruction used for data distillation: sending user instructions and the original text to GPT-4 for compression, and the compressed text is required to complete the user instructions. The highlighted part shows the difference between our method and LLMLingua-2~\cite{llmlingua2} in data collection.}
	\label{fig:prompt}
\end{figure}

Depending on whether task information is utilized, prompt compression methods can be classified as either task-aware or task-agnostic.

Task-aware compression adapts the context based on the downstream task or current query. For example, LongLLMLingua~\cite{Longllmlingua} uses a question-aware coarse-to-fine approach to estimate token information entropy, adjusting it according to the question. Reinforcement Learning (RL)-based methods~\cite{discreteCompress,Boostllm} train models with reward signals from downstream tasks to achieve prompt compression. Soft prompt tuning methods~\cite{promptcontrastive,GIST:NIPS23} typically require fine-tuning for specific tasks, while \cite{recomp:ICLR24} trains a summarization model to compress context based on the query. These task-aware approaches are often tailored to particular tasks and compression ratios, potentially limiting their applicability in real-world scenarios. Compared to these methods, our method can \emph{flexibly switch between task-aware and task-agnostic modes} by simply setting prefix prompts, and can \emph{dynamically select the compression rate}.

In contrast, task-agnostic methods compress prompts without considering specific tasks, making them versatile for various applications and black-box LLMs. However, generating compressed text that generalizes across different tasks is challenging. Information entropy-based metrics are commonly used to prune redundant information~\cite{compresscontext,llmlingua}, with a small language model estimating token importance. While these methods do not require training, they may not optimally capture token importance distributions for specific LLMs and often incur high computational costs. Summarization-based methods are also used~\cite{memorymaze,memGPT} but often exclude essential details and struggle to generalize effectively. Recently, LLMLingua-2~\cite{llmlingua2} proposes a data distillation procedure to derive knowledge from an LLM to compress prompts without losing crucial information.

\section{The Proposed Method}

\begin{figure}
	\centering
	\includegraphics[width=0.9\linewidth]{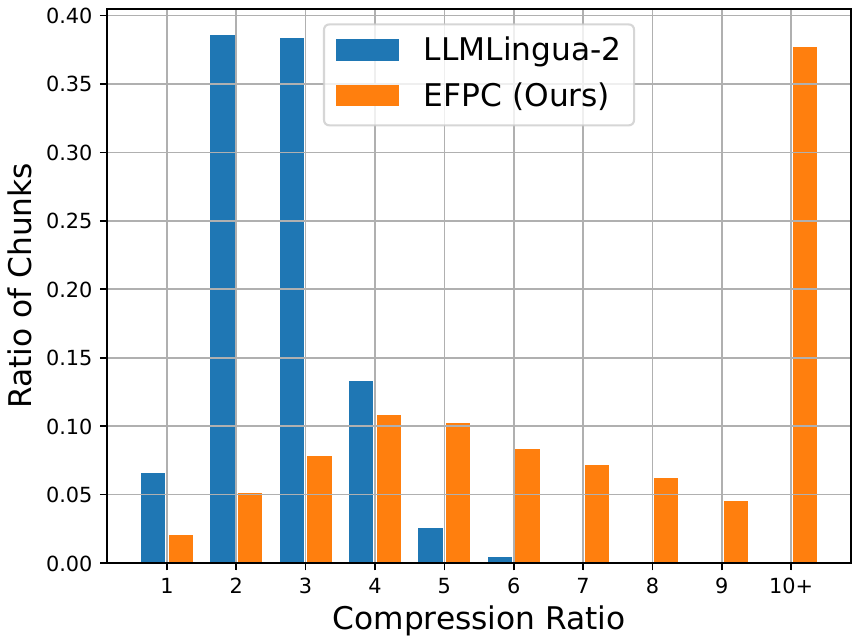}
	\caption{Compression ratio distribution on MeetingBank~\cite{meetingbank}. It shows that our collected dataset achieves a much higher compression ratio (5 times more) compared to LLMLingua-2.}
	\label{fig:histogram}
\end{figure}

\begin{figure*}
	\centering
	\includegraphics[width=\linewidth]{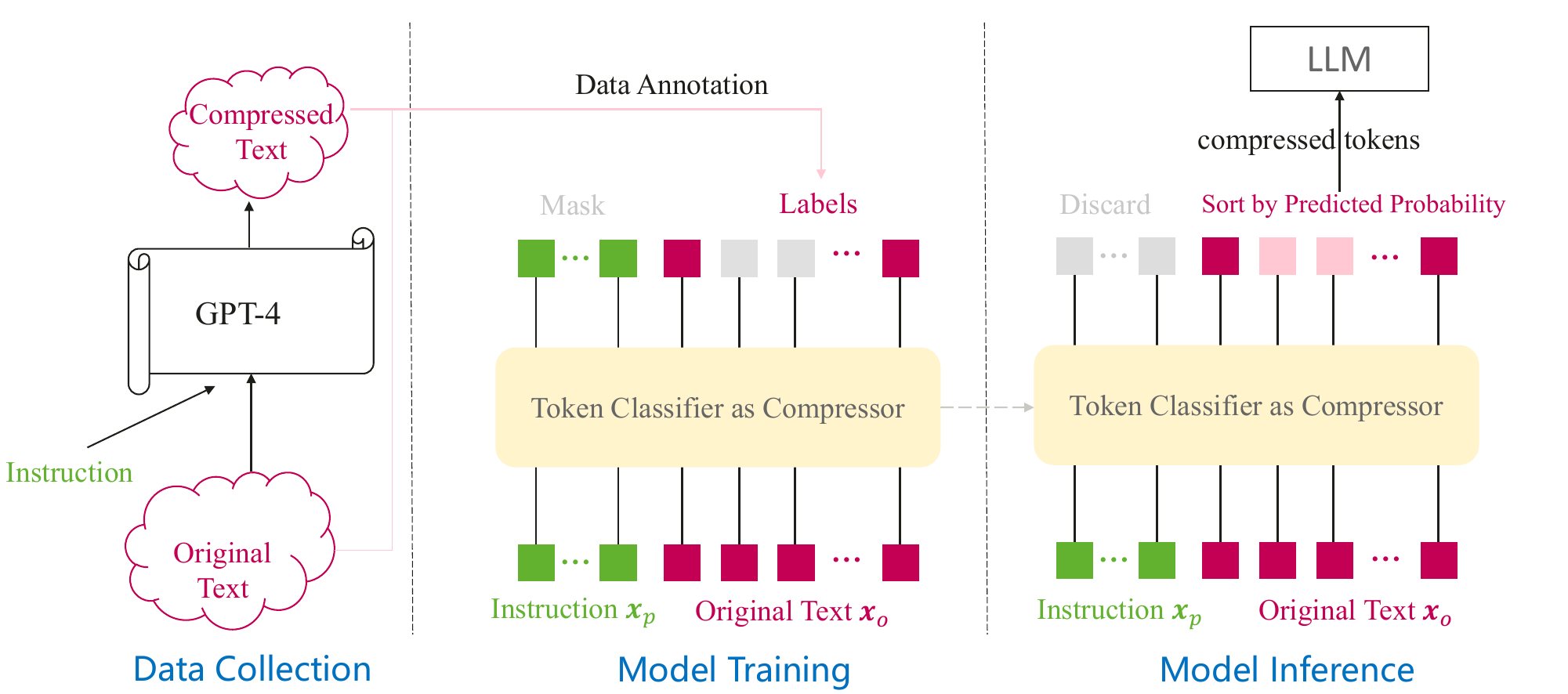}
	\caption{The proposed method. From left to right are the data collection, model training, and inference processes.}
	\label{fig:network}
\end{figure*}

The overall framework of our method is shown in Figure~\ref{fig:network}. We begin with the data collection process in Sec.~\ref{sec:data_collection}, followed by our training and inference algorithm in Sec.~\ref{sec:training}.

\subsection{Instruction-Aware Data Collection}\label{sec:data_collection}

In this paper, we propose an instruction-aware data collection method for prompt compression. As shown in Figure~\ref{fig:prompt}, we send the user instruction together with the original text for compression and ask the LLM to generate compressed text which can complete the user task. Following \citet{llmlingua2}, we segment each long context into chunks, each containing no more than 512 tokens and ending with a period, and instruct GPT-4 to compress each chunk individually.

For fair comparison with LLMLingua-2, we utilize MeetingBank~\cite{meetingbank} as the source dataset. In Figure~\ref{fig:histogram}, we present a histogram of the compression ratio distribution for each chunk. The compression ratio is calculated as follows: LEN(text before compression)/LEN(text after compression), where LEN($\cdot$) denotes the length function. We compare our results with the dataset released by LLMLingua-2\footnote{https://huggingface.co/datasets/microsoft/MeetingBank-LLMCompressed}, which is task-agnostic. As shown in Figure~\ref{fig:histogram}, our instruction-aware data collection method achieves significantly higher compression rates. The average compression ratio of our dataset is 10.9, whereas the LLMLingua-2 dataset has an average ratio of 2.7. This outcome is expected, as fulfilling a specific task often requires only a subset of the original text's information.

For label generation, we first search for the corresponding word in the original prompt for each word in the compressed prompt. If a match is found, the word is assigned a label of 1; otherwise, it is assigned a label of 0.

\subsection{Flexible Token Classification Model}\label{sec:training}

\noindent\textbf{Architecture} Following~\citet{llmlingua2}, we adopt a Transformer encoder as the feature encoder $f_{\theta}$ and a linear classification layer on top. In this paper, we propose instruction-aware training and inference based on a token classification model. The input consists of two parts. The first part is an optional instruction (e.g., a question) composed of \(M\) words, $\boldsymbol{x}_{p}=\{x_i\}_{i=1}^M$, where $M=0$ means that there is no extra instruction input. The second part is the original input text consisting of \(N\) words, $\boldsymbol{x}_{o}=\{x_i\}_{i=M+1}^{M+N}$. The final input concatenates $\boldsymbol{x}_{p}$ before $\boldsymbol{x}_{o}$ and the whole process is:

\begin{align}
	\boldsymbol{h} &= f_{\theta}(\boldsymbol{x}), \\
	p(x_i, \Theta) &= \text{softmax}(W h_i+b),
\end{align}
where $\boldsymbol{x}=[\boldsymbol{x}_{p}, \boldsymbol{x}_{o}]$, $\boldsymbol{h}=\{h_i\}_{i=1}^{M+N}$ denotes feature vectors for all words, $p(x_i, \Theta)\in \mathbb{R}^2$ denotes the probability distribution of labels \{preserve, discard\} for the $i$-th word $x_i$, and $\Theta=\{\theta, W, b\}$ represent all the trainable parameters.

\noindent\textbf{Loss Function} We formulate the prompt compression task as a binary classification problem, using cross entropy (CE) loss for training. Let $\boldsymbol{y}=\{y_i\}_{i=1}^{M+N}\in \{0,1\}$ be the labels corresponding to all words in $\boldsymbol{x}$, where $y_i=0$ means the $i$-th word is discarded, and $y_i=1$ means it is preserved.

\begin{table*}[ht]
	\caption{Data Efficiency of our method. The first row is uncompressed and the second row is LLMLingua-2.}
	\label{tab:incremental}
	\centering
	\small
	\setlength{\tabcolsep}{3pt}
	\begin{tabular}{cc|c| ccc c c}
		\hline
		\multicolumn{2}{c|}{Extra Training Data} & \multirow{2}{*}{\makecell{Compression\\Ratio}} & \multicolumn{5}{c}{LongBench Single-Doc QA} \\
		\cline{4-8}
		Fraction & \# Documents &  & narrativeQA & qasper & multifield\_en  & multifield\_zh & \textbf{AVG} \\
		\hline
		- & - & 1$\times$ &22.87  &41.37  & 52.60 & 60.37 & 44.30 \\
		\hline
		$0\%$ & 0  & \multirow{6}{*}{4.4$\times$}  & 16.01 & 38.75 & 43.48 & 44.16 & 35.16 \\
		$1\%$ & 50 &  & 16.62 & 39.54 & 44.59 & 48.35 & 37.28 \\
		$5\%$ & 250 & & 14.89 & 42.42 & 48.59 & 50.50 & 39.10 \\
		$10\%$ & 500 & & 16.09 & 40.38 & 50.53 & 54.63 & 40.41 \\
		%$20\%$ & 1000 & & 13.18 & 43.57 & 49.50 & 55.03 & 40.32 \\
		$100\%$ & 5000 & & 17.44 & 44.57 & 51.48 & 55.11 & \textbf{42.15} \\
		\hline 
	\end{tabular}
\end{table*}

The labels for the words from $i=M+1$ to $M+N$ are derived directly from the matching results between the original and compressed text. For the instruction part, there are two approaches: one way is to directly set the labels of the prepended part to zero, as it is not intended to be retained in the final output. The corresponding loss function is:

\begin{equation}
	L_{\text{drop}}(\Theta) = \frac{1}{M+N} \sum_{i=1}^{M+N} \text{CE}(y_i; p(x_i,\Theta))\,,
\end{equation}
where $\{y_i\}_{i=1}^M=0$. 

Alternatively, we can ignore this part of the output during training since it is not used when predicting. The corresponding loss function is:

\begin{equation}
	L_{\text{mask}}(\Theta) = \frac{1}{N} \sum_{i=M+1}^{M+N} \text{CE}(y_i; p(x_i,\Theta)).
\end{equation}

For a more intuitive comparison, we also consider the loss function of LLMLingua-2, which is task-agnostic and does not take additional $\boldsymbol{x}_p$ as input ($M=0$). The loss function is defined as follows:
\begin{equation}
	L_{\text{agnostic}}(\Theta) = \frac{1}{N} \sum_{i=1}^{N} \text{CE}(y_i; p(x_i,\Theta)).
\end{equation}

Subsequent results show that both $L_{\text{mask}}$ and $L_{\text{drop}}$ outperform $L_{\text{agnostic}}$, proving the effectiveness of our methods. Notably, $L_{\text{mask}}$ is superior to $L_{\text{drop}}$ because it maintains consistency between tranining and inference, allowing the instruction part to guide the output of the original prompt part $\boldsymbol{x}_o$ without using its prediction probability.

\noindent\textbf{Training Strategy} To equip our model with both task-agnostic and task-aware capabilities, we propose two strategies:

(1) Incremental training. Building on the pre-trained LLMLingua-2 model, we add extra instruction-aware compression data for training, as detailed in Sec.~\ref{sec:data_collection}. Experimental results demonstrate that our method is highly data-efficient and can achieve significant improvements with only a small amount of additional data.

(2) Joint training. We combine task-agnostic data from LLMLingua-2 with our task-aware data. For fair comparisons, we keep the total data amount constant, adjusting only the ratio between the two.

\noindent\textbf{Inference Stage} Our method for compressing the initial prompt $\boldsymbol{x}_o=\{x_i\}_{i=M+1}^{M+N}$ to achieve a target compression ratio of $1/\tau$ involves three main steps. Here, $\tau$ represents the ratio of the number of words in the compressed prompt $\tilde{\boldsymbol{x}}_o$ relative to the number of words in the original prompt $\boldsymbol{x}_o$. First, we determine the target number of words/tokens to be maintained in the compressed 
prompt as $\tilde{N}=\tau N$. Next, the token classification model estimates the probability $p_i$ for each word $x_i$ being assigned the label ``preserve''. Finally, we select the top $\tilde{N}$ words from the original prompt $\boldsymbol{x}_o$ with the highest $p_i$ values, maintaining their original sequence, to construct the compressed prompt $\tilde{\boldsymbol{x}}_o$.

Notably, our model can switch between task-agnostic and task-aware modes during inference by altering the prefix prompt. For instruction-aware tasks, such as question answering, we concatenate the user instruction, $\boldsymbol{x}_p$, with the original text $\boldsymbol{x}_o$ as the input. Conversely, when instruction-agnostic compression is required, we set 
$\boldsymbol{x}_p$ to null.

\section{Experimental Results}

We provide implementation details in Sec.~\ref{sec:details} and evaluate the data efficiency in Sec.~\ref{sec:data_efficiency}. Experiments on benchmark datasets are discussed in 
Sec.~\ref{sec:benchmark}. In Sec.~\ref{sec:ablation}, we conduct ablation studies to analyze the impact of different components integrated into our method. All experiments were conducted using PyTorch on Tesla V100 GPUs.

\begin{table*}[ht]
	\caption{In-domain evaluation of different methods on MeetingBank. Results marked with $^\dagger$ are sourced from ~\citet{llmlingua2}, while those indicated with $^\diamond$ are reproduced using LLama-3.1-8b-instruct to ensure fair comparisons.}
	\label{tab:meetingbank}
	\centering
	\small
	\setlength{\tabcolsep}{3pt}
	\begin{tabular}{c|c| ccc c c|cc}
		\hline
		\multirow{2}{*}{Backbone}& QA & \multicolumn{5}{c|}{Summary} & \multicolumn{2}{c}{Length} \\
		\cline{2-9}
		& F1 Score & BELU & Rouge1 & Rouge2 & RougeL & BERTScore & Tokens & $1/\tau$ \\
		\hline
		Selective-Context~\cite{compresscontext}$^\dagger$ & 66.28 & 10.83 & 39.21 & 18.73 & 27.67 & 84.48 & 1,222 & 2.5$\times$\\
		LLMLingua~\cite{llmlingua}$^\dagger$ & 67.52 & 8.94 & 37.98 & 14.08 & 26.58 & 86.42 & 1,176 & 2.5$\times$ \\
		LLMLingua-2~\cite{llmlingua2}$^\diamond$ & 77.26 &  15.65 & 41.47 & 18.94  & 30.41  & 86.50   & 984 & 3.0$\times$ \\
		EFPC (Ours)$^\diamond$ & 83.35 & 16.71 & 42.29 & 20.63 & 31.75 & 87.49 & 998 & 3.0$\times$
		\\
		\hline 
		Original$^\diamond$ & 84.46 & 17.32 & 43.19 & 22.90  & 33.96 & 88.24  & 3,003 & 1.0$\times$ \\
		\hline 
	\end{tabular}
\end{table*}

\subsection{Implementation Details}\label{sec:details}

\begin{figure}
	\centering
	\includegraphics[width=\linewidth]{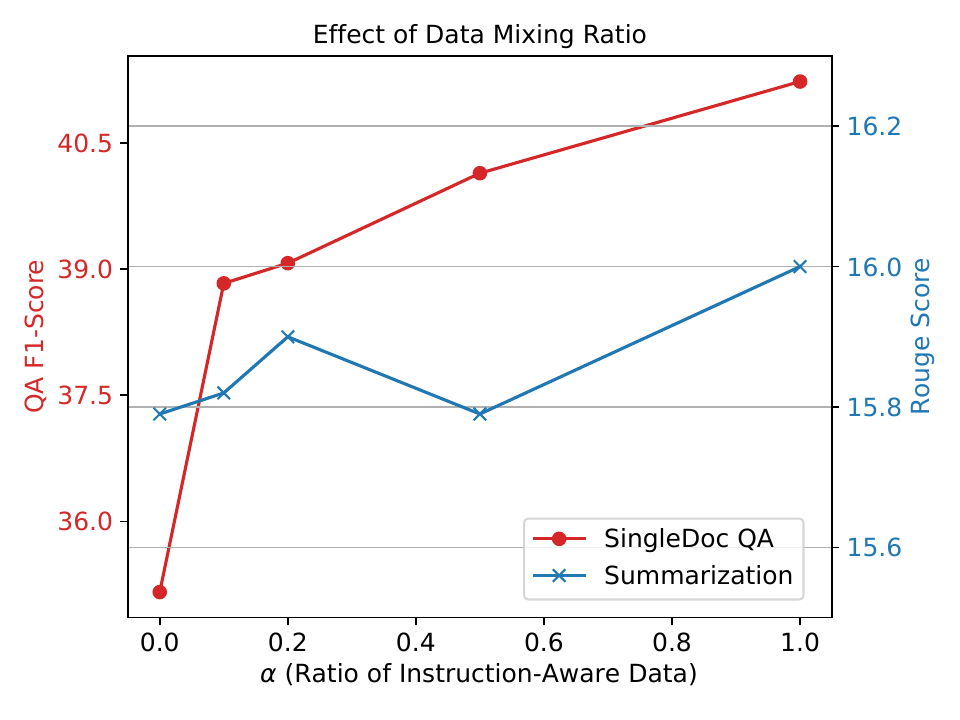}
	\caption{Data Efficiency of our method. We keep the same amount of training data and set target token to 3,000 during compression.}
	\label{fig:joint}
\end{figure}

\begin{table*}[ht]
	\caption{Out-of-domain evaluation on general long-context scenarios. Results with $^\dagger$ come from ~\citet{llmlingua2}}
	\label{tab:longbench}
	\renewcommand{\arraystretch}{0.95}
	\centering
	\small
	\setlength{\tabcolsep}{2pt}
	\begin{tabular}{c|c cccc cccc|ccc}
		\hline
		\multirow{2}{*}{\textbf{Methods}}& \multicolumn{9}{c|}{\textbf{LongBench}} & \multicolumn{3}{c}{\textbf{ZeroSCROLLS}} \\
		\cline{2-13}
		&  Single-Doc & Multi-Doc & Summ. & FewShot & Synth. & Code & \textbf{AVG} & Tokens & $1/\tau$ & \textbf{AVG} & Tokens & $1/\tau$\\
		\hline 
		\multicolumn{13}{c}{\textit{2,000-token constraint}} \\
		\hline
		\multicolumn{13}{l}{\textit{Task(Question)-Agnostic Compression}} \\
		Selective-Context~\citeyearpar{compresscontext}$^\dagger$ & 16.2 & 34.8 & 24.4 & 15.7 & 8.4 & 49.2 & 24.8 & 1,925 & 5$\times$ & 19.4 &  1,865 & 5$\times$ \\
		LLMLingua~\citeyearpar{llmlingua}$^\dagger$ & 22.4 & 32.1&  24.5& 61.2& 10.4 & 56.8 & 34.6 & 1,950 & 5$\times$ & 27.2 & 1,862 & 5$\times$\\
		LLMLingua2~\citeyearpar{llmlingua2}$^\dagger$ & 29.8 & 33.1 & 25.3 & 66.4 & 21.3 & 58.9 & 39.1 & 1,954 & 5$\times$ & 33.4 & 1,898 & 5$\times$\\
		%LLMLingua-2 (repro) & 30.9 & 29.1 & 13.8 & 30.4 & 46.0 & 21.4 &   \\
		%\hline 
		%\multicolumn{13}{l}{\textit{Task(Question)-Aware Compression}} \\
		SBERT~\citeyearpar{sbert}$^\dagger$ & 33.8 & 35.9 & 25.9 & 23.5 & 18.0 & 17.8 & 25.8 & 1,947 & 5$\times$  & 20.5 & 1,773 & 6$\times$  \\
		OpenAI$^\dagger$ & 34.3 & 36.3 & 24.7 & 32.4 & 26.3 & 24.8 & 29.8 & 1,991 & 5$\times$  & 20.6 & 1,784 & 5$\times$ \\
		LongLLMLingua~\citeyearpar{Longllmlingua}$^\dagger$ & 39.0 & 42.2 & 27.4 & 69.3 & 53.8 & 56.6 & 48.0 & 1,809 & 6$\times$  & 32.5 & 1,753 & 6$\times$  \\
		\textbf{EFPC (Ours)} & 41.7 & 42.2 & 25.8 &  67.3 &  27.0 & 57.6 & 43.6 & 1,972 &5$\times$  & 32.7 & 1,877 & 5$\times$ \\
		\hline 
		\multicolumn{13}{c}{\textit{3,000-token constraint}} \\
		\hline
		%\multicolumn{13}{l}{\textit{Task(Question)-Agnostic Compression}} \\
		Selective-Context~\citeyearpar{compresscontext}$^\dagger$ & 23.3 & 39.2 & 25.0 & 23.8 & 27.5 & 53.1 & 32.0 & 3,328 & 3$\times$ & 20.7 & 3,460 & 3$\times$\\
		LLMLingua~\citeyearpar{llmlingua}$^\dagger$ & 31.8 & 37.5 & 26.2 & 67.2 & 8.3 & 53.2 & 37.4 & 3,421 & 3$\times$ & 30.7 & 3,366 & 3$\times$\\
		LLMLingua-2~\citeyearpar{llmlingua2}$^\dagger$ & 35.5 & 38.7 & 26.3 & 69.6 & 21.4 & 62.8 & 42.4 & 3,392 & 3$\times$ & 33.5 & 3,206& 3$\times$  \\
		%\hline 
		%\multicolumn{13}{l}{\textit{Task(Question)-Aware Compression}} \\
		SBERT~\citeyearpar{sbert}$^\dagger$ & 35.3 & 37.4 & 26.7 & 63.4 & 51.0 & 34.5 & 41.4 & 3,399 & 3$\times$ & 24.0 & 3,340 & 3$\times$ \\
		OpenAI$^\dagger$ & 34.5 & 38.6 & 26.8 & 63.4 & 49.6 & 37.6 & 41.7 & 3,421 & 3$\times$ & 22.4 & 3,362 & 3$\times$ \\
		LongLLMLingua~\citeyearpar{Longllmlingua}$^\dagger$ & 40.7 & 46.2 & 27.2 & 70.6  & 53.0 & 55.2 & 48.8 & 3,283 & 3$\times$ & 32.8 & 3,412 & 3$\times$ \\
		\textbf{EFPC (Ours)} & 42.9 & 46.6 & 26.9 & 70.4 & 32.5 & 59.7 & 46.5 & 3,415 &   3$\times$ & 33.9 & 3,327 & 3$\times$ \\
		\hline
		\hline
		Original Prompt$^\dagger$ & 39.7 &38.7 &26.5 &67.0 & 37.8 & 54.2 & 44.0 & 10,295 & - & 34.7 & 9,788 & - \\
		\hline
		Zero-Shot$^\dagger$ & 15.6 & 31.3 & 15.6 & 40.7 & 1.6 & 36.2 & 23.5 & 214 & 48$\times$ & 10.8 & 32 & 306$\times$ \\
		\hline
	\end{tabular}
\end{table*}

\noindent\textbf{Datasets} To ensure a fair comparison with LLMLingua-2~\cite{llmlingua2}, we construct our text compression dataset using training examples from MeetingBank~\cite{meetingbank} as the training set, as illustrated in Sec.~\ref{sec:data_collection}. The compressed prompts are evaluated on two groups of datasets:

(i) \textbf{In-Domain:} We utilize the MeetingBank test set for in-domain evaluation. In addition to the summarization task, we adopt the QA task as outlined in~\cite{llmlingua2}. For both tasks, we used the same evaluation metrics as~\cite{llmlingua2}.

(ii) \textbf{Out-of-Domain:} For long-context scenarios, we use LongBench~\cite{longbench} and Zero-SCROLLS~\cite{zeroscrolls}, employing the same evaluation metrics as LongLLMLingua~\cite{Longllmlingua}. %For reasoning and in-context learning, we use GSM8K (Cobbe et al., 2021), with evaluation metrics consistent with LLMLingua~\cite{llmlingua} and LLMLingua-2~\cite{llmlingua2}.

\noindent\textbf{Training Details} Our approach is implemented using Huggingface’s Transformers and PyTorch 2.0.1 with CUDA-12.2. We use \texttt{multilingual-BERT}~\cite{bert:devlin:NAACL19} as the feature encoder $f_{\theta}$ in our compressor, fine-tuning both models for 10 epochs. We employe the Adam optimizer~\cite{adam:ICLR15} with a learning rate of 1e-5 and a batch size of 10. Unless specified otherwise, due to resource constraints, all reported metrics use \texttt{multilingual-BERT} as the compressor and \texttt{Llama-3.1-8b-instruct}~\cite{llama3} as the target LLM for downstream tasks. For fair comparison with mainstream methods in Table~\ref{tab:longbench}, we employ \texttt{GPT-3.5-Turbo-0613} as the target LLM for downstream tasks, with greedy decoding at a temperature of 0 to ensure enhanced stability across experiments.

\subsection{Data Efficiency of Our Method} \label{sec:data_efficiency}

Before delving into detailed experimental results, it is essential to highlight our method's data efficiency. As discussed in Sec.~\ref{sec:training}, this is evaluated via two primary approaches: incremental training with small additional amounts of data and joint training with fixed data volume but varying proportions of instruction-aware data. As noted in Section~\ref{sec:data_collection}, our collected data is instruction-aware (see Figure~\ref{fig:prompt}), while instruction-agnostic data comes from~\citet{llmlingua2}.

In Table~\ref{tab:incremental}, we extend the training of the pretrained LLMLingua-2 model by adding a small amount of extra data. Different dataset ratios are employed to create subsets with a target token limit of 3,000 for compression, achieving an effective compression rate of approximately 4.4. As shown in Table~\ref{tab:incremental}, our EFPC exhibits remarkable data efficiency. With just a 1\% increment in training data (around 50 additional documents from MeetingBank~\cite{meetingbank}), we see a 6\% improvement in the F1-score on the LongBench single-doc QA benchmark. When the additional data ratio increases to 10\%, the improvement surges to 14.9\%.

\begin{figure*}[t]
	\centering
	\includegraphics[width=\linewidth]{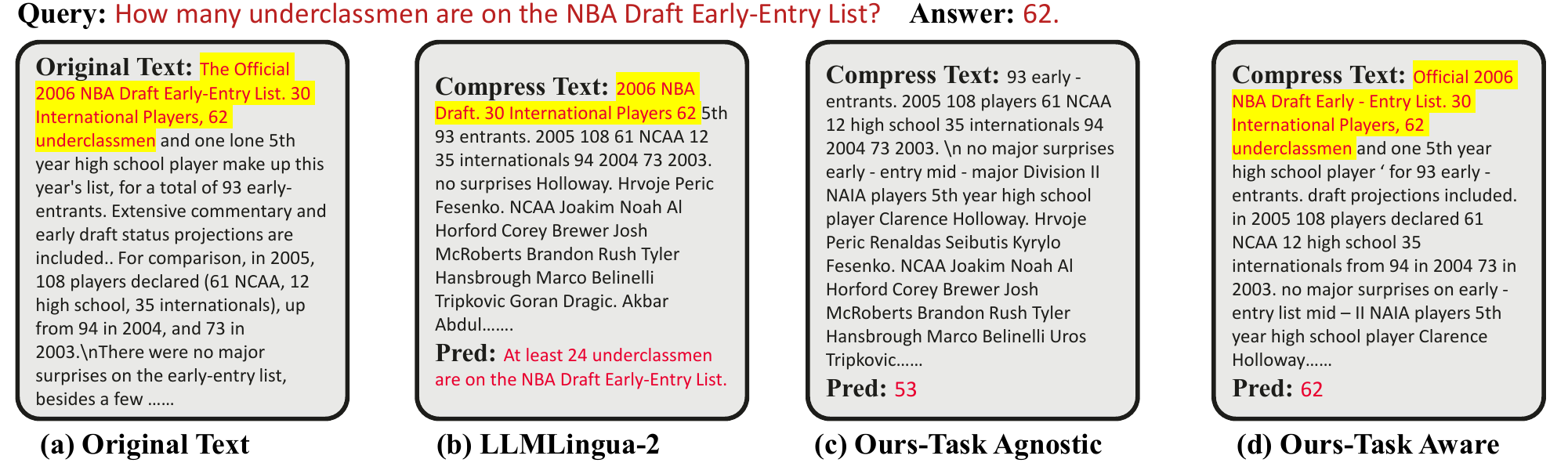}
	\caption{Comparison of compressed text by different methods. (a) Original text. (b) LLMLingua-2. (c) Our method without task-aware inference. (d) Our method with task-aware inference. It can be observed that our method is able to retain more information relevant to the problem, thereby enabling the LLM to generate correct answers.}
	\label{fig:compress_demo}
\end{figure*}

\begin{table*}[t]
	\caption{Ablation Study of task-aware design. For fair comparison, we use 3000-token constraint here.}
	\label{tab:instruction_ablation}
	\renewcommand{\arraystretch}{0.9}
	\centering
	\small
	\begin{tabular}{c|c|c ccc c c}
		\hline
		\multirow{2}{*}{\makecell{Task-Aware\\Training?}}&\multirow{2}{*}{\makecell{Task-Aware\\Inference?}} & \multicolumn{5}{c}{LongBench} \\
		\cline{3-7}
		& & Single-Doc & Multi-Doc & Summ.  & Few-Shot & \textbf{AVG} \\
		\hline
		$\times$ & $\times$ & 35.16 & 31.13 & 15.79 & 33.88 & 28.99 \\
		$\times$ & \checkmark & 36.51 & 31.98 & 15.70 & 34.44 & 29.66 \\
		\checkmark & $\times$ & 35.64 & 31.99 & 15.97 & 37.01 & 30.15 \\
		\checkmark & \checkmark & \textbf{42.15} & \textbf{36.59} & \textbf{16.12} & \textbf{38.99} & \textbf{33.46} \\
		\hline 
	\end{tabular}
\end{table*}

In Figure~\ref{fig:joint}, we keep the total training data volume constant while varying the ratio of instruction-aware data $\alpha$ (with the ratio of instruction-agnostic data being $1-\alpha$). As the proportion of our collected data increased, QA tasks consistently improve, while summarization tasks remain stable.

Experimental results in Table~\ref{tab:incremental} and Figure~\ref{fig:joint} validate our method's effectiveness and data efficiency. Our approach achieves significant performance improvements with minimal additional data, highlighting its potential for efficient data utilization in various applications.

\subsection{Experiments on Benchmark Datasets}\label{sec:benchmark}

\noindent\textbf{Results on In-Domain Benchmark} In Table~\ref{tab:meetingbank}, our proposed method surpasses strong baselines on MeetingBank. Despite the fact that our compressors are much maller than the LLaMa-2-7B~\cite{llama2} used in the baselines, our approach achieves significantly better performance on both the QA and Summary tasks, and nearly matches the original prompt's performance. This demonstrates the effectiveness of our constructed dataset and the benefits of optimizing the compression model using prompt compression knowledge.

\noindent\textbf{Results on Out-of-Domain Benchmark} As shown in Table~\ref{tab:longbench}, EFPC outperforms LLMLingua-2, which is based on the same lightweight model, by 39.9\% on single-doc dataset and 20.8\% on multi-doc dataset at a 5$\times$ compression ratio. Our method also achieves comparable results to LongLLMLingua~\cite{Longllmlingua}, which is based on a much heavier decoder structure. Notably, LongLLMLingua incurs higher overhead in terms of memory, latency, and power consumption during inference, as will be shown in Sec.~\ref{sec:latency}.

Moreover, as the compression ratio increases from 3$\times$ to 5$\times$, LLMLingua2 exhibits a performance drop of 16.1\% on single-doc QA, whereas our method only experiences a decline of 2.8\%. This indicates that our approach is more effective at higher compression ratios, showcasing a stronger ability to preserve essential information.

\begin{table}
	\caption{Ablation study of loss function. We use 50\% extra data with 3000-token constraint here.}
	\label{tab:loss_ablation}
	\renewcommand{\arraystretch}{0.9}
	\centering
	\small
	\setlength{\tabcolsep}{1pt}
	\begin{tabular}{c| ccc c c}
		\hline
		\multirow{2}{*}{Loss Function} & \multicolumn{5}{c}{LongBench} \\
		\cline{2-6}
		& Single-Doc & Multi-Doc & Summ.  & Few-Shot & \textbf{AVG} \\
		\hline
		$L_{\text{agnostic}}$ & 35.16 & 31.13 & 15.79 & 33.88 & 28.99\\
		$L_{\text{drop}}$ & 41.40 & 35.78 & 15.98 & 36.14 & 32.33 \\
		$L_{\text{mask}}$ (Ours) & \textbf{42.15} & \textbf{36.59} & \textbf{16.12} & \textbf{38.99} & \textbf{33.46} \\
		\hline 
	\end{tabular}
\end{table}

\begin{table*}
	\caption{Evaluation with Mistral-7B as the Target LLM on MeetingBank and LongBench single-doc QA task. We report Rouge1~\cite{rouge} for summary. $^\dagger$: Results from~\citet{llmlingua2}.}
	\label{tab:mistral}
	\renewcommand{\arraystretch}{0.95}
	\centering
	\small
	\setlength{\tabcolsep}{2.5pt}
	\begin{tabular}{c|cccc| cccccc}
		\hline
		\multirow{2}{*}{\textbf{Methods}}& \multicolumn{4}{c|}{\textbf{MeetingBank}} & \multicolumn{6}{c}{\textbf{LongBench Single-Doc}} \\
		\cline{2-11}
		& QA & Summ. & Tokens & 1/$\tau$ & 2,000-token cons. & Tokens & 1/$\tau$ & 3,000-token cons. & Tokens & 1/$\tau$  \\
		\hline
		Selective-Context~\citeyearpar{compresscontext}$^\dagger$ & 58.13 & 26.84 & 1,222 & 2.5$\times$ & 22.0 & 2,038 & 7.1$\times$ & 26.0 & 3,075 & 4.7$\times$ \\
		LLMLingua~\citeyearpar{llmlingua}$^\dagger$ & 50.45 & 23.63 & 1,176 & 2.5$\times$ & 19.5 & 2,054 & 7.1$\times$ & 20.8 & 3,076 & 4.7$\times$ \\
		LLMLingua-2~\citeyearpar{llmlingua2}$^\dagger$ & 75.97 & 29.93 & 984 & 3.0$\times$ & 25.3 & 1,949 & 7.4$\times$ & 27.9 & 2,888 & 5.0$\times$\\
		LLMLingua-2-Large$^\dagger$ & 76.22 & 30.18 & 970 & 3.0$\times$ & 26.8 & 1,967 & 7.4$\times$ & 27.3 & 2,853 & 5.1$\times$\\
		% LLMLingua-2 (Repro) & \\
		\textbf{EFPC (Ours)} & 80.51  &29.72 & 992 & 3.0$\times$ & 29.7 & 1,994 & 7.3$\times$ & 30.2 & 2,850 & 5.1$\times$ \\
		\textbf{EFPC-Large (Ours)} & \textbf{81.63} &\textbf{30.43} &985 & 3.0$\times$ & \textbf{31.8} & 2,002 & 7.2$\times$ & \textbf{31.9} & 2,913& 5.0$\times$  \\
		% EFPC-large (Ours) & \\
		\hline 
		Original Prompt$^\dagger$ & 66.95 & 26.26 & 3,003 & - & 24.5 & 14,511 & - & 24.5 & 14,511 & - \\
		\hline
	\end{tabular}
	\vspace{-6pt}
\end{table*}

\subsection{Ablation Study}\label{sec:ablation}
In this section, we conduct ablation experiments to evaluate the impact of different components in our proposed method (Table~\ref{tab:instruction_ablation}) and analyze the effect of different loss functions (Table~\ref{tab:loss_ablation}).

\noindent\textbf{Task-aware Training and Inference} We examine the effects of task-aware training and inference. Task-aware inference involves prepending the task instruction to the input (e.g., a query for QA datasets) during inference. Table~\ref{tab:instruction_ablation} shows that without task-aware training or inference (first row), the method degrades to LLMLingua-2. When both are used (last row), it represents our EFPC method. Results indicate that task-aware inference consistently improves accuracy, regardless of whether task-aware training is used. Moreover, the combination of both yields the most significant benefits. For example, using task-aware inference without task-aware training improves relative accuracy by 3.84\% and 2.73\% for single-doc and multi-doc QA datasets, respectively. When both are employed (our EFPC), relative improvements increase to 10.95\% and 10.24\%.

In Figure~\ref{fig:compress_demo}, we compare the compressed texts using different methods (with the 2,000-token constraint for all). Both columns (b) and (c), which do not employ task-aware inference, result in compressed texts that lose critical information required to answer user questions. For instance, column (b) omits "underclassmen". In contrast, our method in column (d) retains all essential information. It is worth noting that the same compression ratio is used in columns (b), (c), and (d). Notably, column (d) demonstrates a variable compression pattern: key regions are compressed less, while non-essential regions are compressed more.

\noindent\textbf{Loss Function} In Table~\ref{tab:loss_ablation}, we compare different loss functions from Sec.~\ref{sec:training}. Both $L_{\text{mask}}$ and $L_{\text{drop}}$ outperform $L_{\text{agnostic}}$, which is equivalent to LLMLingua-2~\cite{llmlingua2}. Comparing \(L_\text{drop}\) and \(L_\text{mask}\), we see that ignoring the labels and predictions of the instruction part during training yields better results. This is because it ensures consistency between the training and inference stages. 

In summary, the experiments in Table~\ref{tab:instruction_ablation} and Figure~\ref{fig:compress_demo} show that both task-aware training and inference are effective. Combining both (ensuring consistency between training and inference) yields the highest gains. Additionally, Table~\ref{tab:loss_ablation} demonstrates the effectiveness of the proposed $L_{\text{mask}}$.

\subsection{Generalizability}
In this section, we validate our method's generalizability across different model architectures, including prompt compression model and LLM used for generating responses. Previously, we tested primarily on the bert-base model, but here we extend our evaluation to \texttt{xlm-roberta-large}~\cite{xlmroberta} and use \texttt{Mistral-7b-v0.1}~\cite{mistral} as the target LLM.

As shown in Table~\ref{tab:mistral}, on the in-domain dataset MeetingBank QA, our method achieve absolute improvements of 4.5\% and 5.4\% over the best baseline method, LLMLingua-2, when using \texttt{multilingual-BERT} and \texttt{xlm-roberta-large}, respectively. This trend is consistent in out-of-domain evaluations as well. The results in Table~\ref{tab:mistral} demonstrate that our method is robust across various backbones, tasks, domains and target LLMs.

%We also verify the effectiveness of the proposed method on the GSM8K dataset for reasoning and in-context learning. As shown in Table~\ref{tab:gsm8k}, 

%\begin{table}
%	\caption{Results on GSM-8k dataset. $^\dagger$: Results from LLMLingua-2~\cite{llmlingua2}.}
%	\label{tab:gsm8k}
%\renewcommand{\arraystretch}{0.9}
%	\centering
%	\small
%	\setlength{\tabcolsep}{3pt}
%	\begin{tabular}{c| ccc |ccc}
	%		\hline
	%		\multirow{2}{*}{Method} & \multicolumn{3}{c|}{1-shot constraint} &  \multicolumn{3}{c}{half-shot constraint}  \\
	%		\cline{2-7}
	%		& EM & Tokens & $1/\tau$ & EM & Tokens & $1/\tau$ \\
	%		\hline
	%		Selective-Context$^\dagger$ & 53.98 & 452 & 5$\times$ & 52.99 & 218 & 11$\times$ \\
	%		LLMLingua$^\dagger$ & 79.08 & 446 & 5$\times$ & 77.41 & 171 & 14$\times$\\
	%		LLMLingua2$^\dagger$ & 78.92 & 437 & 5$\times$ & 77.48 & 161 & 14$\times$ \\
	%		EFPC (Ours) &  \\
	%		\hline 
	%		Full-Shot & 78.85 & 2,366 & - & 78.85 & 2,366 & -\\
	%		Zero-Shot & 48.75 & 11 & 215$\times$ & 48.75 & 11 & 215$\times$ \\
	%		\hline 
	%	\end{tabular}
%\end{table}

\begin{table}
	\caption{Efficiency comparison on MeetingBank. $^\diamond$: We re-evaluate these metrics on Tesla V100 GPU. $^\dagger$: Results from~\citet{llmlingua2}.}
	\label{tab:efficiency}
	\centering
	\small
	\renewcommand{\arraystretch}{0.9}
	\setlength{\tabcolsep}{3pt}
	\begin{tabular}{c |ccc c|c}
		\hline
		$1/\tau$ & 1$\times$ & 2$\times$ & 3$\times$ & 5$\times$  & GPU \\
		\cline{1-5}
		End2End w/o Compression$^\diamond$ & \multicolumn{4}{c|}{15.8} & Mem. \\
		End2End w/ EFPC$^\diamond$ & - & 10.0 & 8.8 & 6.7 & (GB)\\
		\hline
		Selective-Context$^\dagger$ & - & 15.9 & 15.6 & 15.5 & 26.5  \\
		LLMLingua$^\dagger$ & - & 2.9 & 2.1& 1.5 & 16.6 \\
		LLMLingua-2$^\dagger$ &-  &0.5 &0.4 & 0.4 & 2.1 \\
		LLMLingua-2$^\diamond$ &- & 0.4 & 0.4 & 0.4 & 2.1\\
		EFPC$^\diamond$  &- & 0.4 & 0.4 & 0.4 & 2.1 \\
		\hline 
	\end{tabular}
	\vspace{-6pt}
\end{table}

\subsection{Latency Comparison}\label{sec:latency}

Table~\ref{tab:efficiency} presents the latency and GPU memory usage of different methods with various compression rates. Sharing the same lightweight architecture, our method 
shows nearly the same latency as LLM-Lingua2. Compared to other compression techniques, EFPC results in significantly lower computational overhead and achieves an end-to-end speedup of 1.6$\times$ to 2.4$\times$. Additionally, our approach can reduce GPU memory costs by up to 8$\times$, decreasing the demand for hardware resources.

%\begin{table}
%	\caption{Comparisons of different method.}
%	\label{tab:acc_efficiency}
%\renewcommand{\arraystretch}{0.9}
%	\centering
%\small
%\setlength{\tabcolsep}{3pt}
%	\begin{tabular}{c| ccc c c}
	%		\hline
	%		\multirow{2}{*}{Inference Speed} & \multicolumn{5}{c}{LongBench} \\
	%		\cline{2-6}
	%		& SingleDoc & MultiDoc & Summ.  & Synth. & \textbf{AVG} \\
	%		\hline
	
	%		\hline
	%	\end{tabular}
%\end{table}

\section{Conclusions}\label{sec:future}

In this paper, we proposed an efficient and flexible prompt compression method EFPC, for improving data, training, and inference efficiency. We identified the challenges faced by existing methods and addressed them accordingly. With only a single model and once training, our method supports both task-aware and task-agnostic modes, allowing flexible switching between both modes as needed. Extensive experiments across various tasks and domains demonstrate that our approach significantly outperforms other baseline methods in terms of performance and compression latency.

\clearpage

\section*{Limitations}

While our proposed Efficient and Flexible Prompt Compression (EFPC) method demonstrates significant improvements in efficiency and performance, several limitations and areas for future work remain noteworthy.

First, although we emphasize the data efficiency of our method, the training data for our experiments was exclusively sourced from MeetingBank. This was done to ensure a fair comparison with existing methods. While this limitation allowed us to demonstrate notable improvements and generalizability on out-of-domain datasets, we did not explore the potential gains that could be achieved with larger-scale training datasets. Future research could investigate the impact of training with more extensive datasets, which may reveal additional performance enhancements and insights into the scalability of our method.

Second, due to resource constraints, our experiments utilized an 8-billion parameter large language model (LLM). We did not systematically explore how our prompt compression method performs across different sizes of LLMs. Intuitively, larger LLMs might handle prompt compression more effectively, potentially resulting in smaller accuracy losses at higher compression rates. Exploring the sensitivity of our method to various LLM sizes could provide a deeper understanding of its adaptability and robustness, offering valuable guidance for future deployments in diverse computing environments.

Lastly, EFPC, while designed for flexibility and efficiency, may still face challenges when applied to extremely high-compression scenarios or highly specialized tasks that were not covered in our experimental setup. Future research should explore these scenarios to further enhance the robustness and versatility of our approach.

In summary, while EFPC represents a significant step forward in the realm of prompt compression, addressing these limitations could unlock further potential and showcase its applicability across an even broader range of tasks and models. We encourage future work to explore these dimensions to fully realize the capabilities of prompt compression in advancing natural language processing.

\bibliography{custom}
%\appendix

%\section{Example Appendix}
%\label{sec:appendix}

%This is an appendix.

\end{document}